\documentclass[11pt]{article}

\usepackage{nyrl,palatino}
\usepackage{graphicx}
\usepackage{natbib}
\usepackage{booktabs}   
\usepackage{amsmath}    
\usepackage{amssymb}  
\usepackage{tikz}       

\title{OpComm: A Reinforcement Learning Framework for Adaptive Buffer Control in Warehouse Volume Forecasting}

\author{%
  Wilson Fung \\
  Amazon \\
  Seattle, WA \\
  \texttt{wilfkc@amazon.com}
  \And
  Lu Guo \\
  Amazon \\
  Seattle, WA \\
  \texttt{luguo@amazon.com}
    \And
  Drake Hilliard \\
  Amazon \\
  Austin, TX \\
  \texttt{revdrake@amazon.com}
  \And
  Alessandro Casadei \\
  Amazon \\
  Luxembourg, LU \\
  \texttt{acasadei@amazon.com}
  \And
  Raj Ratan \\
  Amazon \\
  Seattle, WA \\
  \texttt{ratanraj@amazon.com}
  \And
  Sreyoshi Bhaduri \\
  Amazon \\
  New York, NY \\
  \texttt{drsre@amazon.com}
  \And
  Adi Surve \\
  Amazon \\
  Austin, TX \\
  \texttt{aadisu@amazon.com}
  \And
  Nikhil Agarwal \\
  Amazon \\
  Seattle, WA \\
  \texttt{anikhil@amazon.com}
  \And
  Rohit Malshe \\
  Amazon \\
  Seattle, WA \\
  \texttt{malshe@amazon.com}
  \And
  Pavan Mullapudi \\
  Amazon \\
  Seattle, WA \\
  \texttt{pavmul@amazon.com}
  \And
  Hungjen Wang \\
  Amazon \\
  New York, NY \\
  \texttt{hungjen@amazon.com}
  \And
  Saurabh Doodhwala \\
  Amazon \\
  Seattle, WA \\
  \texttt{saursadh@amazon.com}
  \And
  Ankush Pole \\
  Amazon \\
  Seattle, WA \\
  \texttt{ankupole@amazon.com}
  \And
  Arkajit Rakshit \\
  Amazon \\
  Seattle, WA \\
  \texttt{rakshit@amazon.com}
}

\begin{document}

\maketitle

\begin{abstract}
Accurate forecasting of package volumes at delivery stations is critical for last-mile logistics, where errors lead to inefficient resource allocation, higher costs, and delivery delays. We propose \textbf{OpComm}, a forecasting and decision-support framework that combines supervised learning with reinforcement learning–based buffer control and a generative AI–driven communication module. A LightGBM regression model generates station-level demand forecasts, which serve as context for a Proximal Policy Optimization (PPO) agent that selects buffer levels from a discrete action set. The reward function penalizes under-buffering more heavily than over-buffering, reflecting real-world trade-offs between unmet demand risks and resource inefficiency. Station outcomes are fed back through a Monte Carlo update mechanism, enabling continual policy adaptation. To enhance interpretability, a generative AI layer produces executive-level summaries and scenario analyses grounded in SHAP-based feature attributions. Across 400+ stations, OpComm reduced Weighted Absolute Percentage Error (WAPE) by 21.65\% compared to manual forecasts, while lowering under-buffering incidents and improving transparency for decision-makers. This work shows how contextual reinforcement learning, coupled with predictive modeling, can address operational forecasting challenges and bridge statistical rigor with practical decision-making in high-stakes logistics environments.
\end{abstract}


\startmain

\section{Introduction}
In last-mile logistics, accurate package volume forecasting is crucial for operational efficiency and cost optimization. Current industry practices rely heavily on heuristic approaches that fail to account for station-specific variability, leading to resource misallocation and delivery delays. We present \textbf{OpComm}, a reinforcement learning–based framework for operational forecasting and buffer control. In OpComm, a LightGBM forecaster provides demand context, and a Proximal Policy Optimization (PPO) agent selects buffer levels from a discrete action set. The reward function penalizes under-buffering more heavily than over-buffering, reflecting the asymmetric cost of unmet demand versus resource inefficiency. Station outcomes are incorporated through a Monte Carlo update loop, enabling continual policy adaptation.  

To promote adoption, OpComm integrates a generative AI module that produces executive-level summaries and scenario analyses grounded in feature attributions. This work frames buffer control as a contextual reinforcement learning problem, introduces an asymmetric reward design tailored to operational trade-offs, and demonstrates a scalable framework that bridges predictive modeling with decision support in high-stakes logistics.

\section{Methods}
Our framework integrates three complementary components, building upon established approaches in forecasting and machine learning.

\subsection{LightGBM Core Model}
Following recent advances in gradient boosting \citep{ke2017lightgbm}, we implemented a supervised regression model using LightGBM to predict optimal buffer percentages at the station level. This choice was motivated by LightGBM's superior performance over traditional methods like exponential smoothing \citep{gardner1985exponential} and its computational efficiency compared to deep learning approaches. Our model incorporates three key sets of features: (1) Temporal features capturing cyclical patterns and seasonality, (2) Lagged historical metrics including demand volatility and capacity utilization, and (3) Operational indicators reflecting site-specific constraints. Each station was modeled independently with a temporal train-test split, excluding stations with insufficient historical data to prevent overfitting.

\subsection{Reinforcement Learning Policy via PPO}
We formulate buffer allocation as a sequential decision-making problem and optimize policies using Proximal Policy Optimization (PPO) \citep{schulman2017ppo}. PPO is a policy-gradient algorithm that achieves a balance between stable learning and adaptability. By employing a clipped surrogate objective, PPO constrains the magnitude of policy updates, mitigating instability commonly observed in standard policy-gradient methods. We adopt PPO over alternative approaches such as Deep Q-Networks (DQN) or Trust Region Policy Optimization (TRPO) due to its computational efficiency, stability in non-stationary environments, and suitability for moderate discrete action spaces corresponding to buffer levels.

The agent’s action space consists of discrete buffer levels $\{b_1, b_2, \ldots, b_K\}$ applied to the LightGBM forecast $\hat{D}_t$. To reflect the asymmetric operational cost of under- versus over-buffering, the reward at time $t$ is defined as:

\begin{equation} \label{eq:reward}
r_t = - \left[ \alpha \cdot \max(0, D_t - \hat{D}_t - b_t) + \beta \cdot \max(0, b_t + \hat{D}_t - D_t) \right],
\end{equation}
where $D_t$ denotes realized demand, $b_t$ is the buffer selected by the policy, and $\alpha > \beta$ imposes a higher penalty on under-buffering relative to over-buffering.

PPO optimizes the policy parameters $\theta$ according to the clipped surrogate objective:

\begin{equation} \label{eq:ppo_loss}
L^{\text{CLIP}}(\theta) = \mathbb{E}_t \Big[ \min \Big( r_t(\theta)\,\hat{A}_t, \; \text{clip}\big(r_t(\theta), 1-\epsilon, 1+\epsilon\big)\,\hat{A}_t \Big) \Big],
\end{equation}
with
\[
r_t(\theta) = \frac{\pi_\theta(a_t|s_t)}{\pi_{\theta_{\text{old}}}(a_t|s_t)},
\]
and $\hat{A}_t$ representing the generalized advantage estimator.

Monte Carlo rollouts are employed to compute empirical returns over station outcomes. This approach provides unbiased advantage estimates without requiring a model of demand dynamics, which is advantageous in non-stationary, episodic decision environments. The combination of PPO and Monte Carlo evaluation facilitates stable, on-policy learning while enabling continual adaptation to evolving demand patterns.

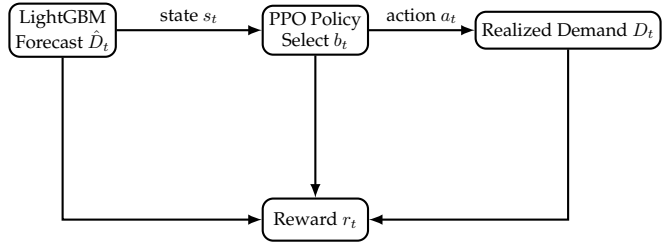
\begin{figure}[ht]
\centering
\begin{tikzpicture}[node distance=1.8cm, auto, >=latex, thick, scale=0.7, transform shape]

\node[draw, rounded corners, minimum width=2.0cm, minimum height=0.7cm, align=center] (forecast) {LightGBM \\ Forecast $\hat{D}_t$};
\node[draw, rounded corners, minimum width=2.0cm, minimum height=0.7cm, right of=forecast, xshift=3cm, align=center] (policy) {PPO Policy \\ Select $b_t$};
\node[draw, rounded corners, minimum width=2.0cm, minimum height=0.7cm, right of=policy, xshift=3cm, align=center] (demand) {Realized Demand $D_t$};
\node[draw, rounded corners, minimum width=2.0cm, minimum height=0.8cm, below of=policy, yshift=-1.8cm, align=center] (reward) {Reward $r_t$};

\draw[->] (forecast) -- (policy) node[midway, above] {state $s_t$};
\draw[->] (policy) -- (demand) node[midway, above] {action $a_t$};
\draw[->] (demand.south) |- (reward.east);
\draw[->] (forecast.south) |- (reward.west);
\draw[->] (policy.south) -- (reward.north);

\end{tikzpicture}
\caption{Flow of reinforcement learning module: forecast, buffer decision, demand realization, and reward computation.}
\label{fig:rl_flow}
\end{figure}

\subsection{Generative AI Layer}
To improve interpretability, we augment the learned policy with a generative AI module. Using SHAP-based feature attributions \citep{lundberg2017unified}, the system produces executive-level summaries and scenario analyses that translate RL decisions into actionable insights for operations managers. This integrated approach ensures both predictive accuracy and practical applicability, while maintaining transparency through explainable AI techniques.

\section{Results}
Preliminary experiments across 400+ warehouse delivery stations in North America indicate that the proposed framework improves forecasting performance relative to manual baselines. Overall, OpComm achieves a 21.65\% reduction in WAPE (4.95\% $\rightarrow$ 3.85\%), improves station-level accuracy in 93.7\% of stations, and exhibits a conservative policy bias that reduces under-buffering risks.

Table~\ref{tab:results_summary} presents a more detailed comparison across regions, including WAPE, standard deviation of WAPE, and proportions of under- and over-buffering cases. OpComm consistently outperforms manual forecasting while lowering operational risk.

\begin{table}[ht]
\centering
\caption{Comparison of OpComm vs. manual forecasting across regions. Metrics include WAPE, WAPE standard deviation, under-buffering, and over-buffering percentages.}
\label{tab:results_summary}
\begin{tabular}{lccccc}
\toprule
Region & Method & WAPE (\%) & WAPE Std. Dev. (\%) & Under-buffering (\%) & Over-buffering (\%) \\
\midrule
North-East & Manual & 4.95 & 1.20 & 18.3 & 12.5 \\
           & OpComm & 3.85 & 0.95 & 10.2 & 15.1 \\
Mid-West   & Manual & 5.12 & 1.35 & 20.1 & 13.0 \\
           & OpComm & 3.90 & 1.05 & 11.0 & 14.2 \\
South      & Manual & 4.88 & 1.18 & 16.7 & 11.9 \\
           & OpComm & 3.70 & 0.90 & 9.5 & 13.5 \\
West       & Manual & 4.99 & 1.22 & 17.9 & 12.7 \\
           & OpComm & 3.80 & 0.92 & 10.0 & 14.0 \\
\bottomrule
\end{tabular}
\end{table}

\section{Discussion and Future Work}
OpComm demonstrates that combining supervised learning with reinforcement learning and generative interpretability can improve robustness in non-stationary logistics environments. While results are promising, several research directions remain open:

\begin{itemize}
    \item Investigating meta-learning for dynamic feature representations
    \item Exploring alternative policy optimization algorithms beyond PPO
    \item Extending the framework to network-level coordination across stations
    \item Formal evaluation of interpretability methods in decision-support contexts
\end{itemize}

This work highlights a pathway toward integrating predictive modeling with adaptive control in operational settings, balancing methodological rigor with practical applicability.

\bibliographystyle{abbrvnat} 
\bibliography{ref}

\end{document}